\documentclass[12pt, oneside]{article}   	
\usepackage{geometry}                		
\usepackage{graphicx}				
\usepackage{amssymb}
\usepackage{url}

\title{Detecting Suspicious events in \\ Fast Information Flows}
\author{Kristiaan Pelckmans, Moustafa Aboushady, Andreas Br\"osemyr}

\begin{document}
\maketitle

\section*{Summary}

We describe a computational feather-light and intuitive, yet provably efficient algorithm, named HALFADO.
HALFADO is designed for detecting suspicious events in a high-frequency stream of complex entries, 
based on a relatively small number of examples of human judgement.
Operating a sufficiently accurate detection system is vital for {\em assisting} teams of human experts in many different areas of the modern digital society.
These systems have intrinsically a far-reaching normative effect, and public knowledge of the workings of such technology should be a human right.

On a conceptual level, the present approach extends one of the most classical learning algorithms for classification, inheriting its theoretical properties.
It however works in a semi-supervised way integrating human and computational intelligence.
On a practical level, this algorithm transcends existing approaches (expert systems) by managing and boosting their performance into a single global detector.

We illustrate HALFADO's efficacy on two challenging applications: 
(1) for detecting {\em hate speech} messages in a flow of text messages gathered from a social media platform, and 
(2) for a Transaction Monitoring System (TMS) in FinTech detecting fraudulent transactions in a stream of financial transactions.

This algorithm illustrates that - contrary to popular belief - advanced methods of machine learning need not require neither advanced levels of computation power
nor expensive annotation efforts.

\newpage

The ability to detect {\em suspicious} events is arguably the most fundamental property of intelligent beings.
The expression of this capability as a machine-executable recipe (an {\em algorithm})  is however challenging.
Approaches based on supervised learning, model identification or estimation, or density estimation are often 
governed by information-theoretical, labeling and/or computational limitations,
hampering the ability to handle effectively the present complex and high-frequency information flows.

The conventional approach is to divide this problem into a {\em training phase} and a subsequent {\em testing} (and/or deployment) phase:
asking the fundamental question at what instance to change gears in order to perform overall well.
Moreover, training requires often  unrealistic annotation efforts.
Approaches based on existing un- or semi-supervised machine learning 
rely heavily on assumptions that might or might not hold in the present case, rendering the result conditional.
Practical considerations constrain the setting even further:
typically, one has only access to finite memory and modest computation power to do this,
we don't have a budget to annotate ('label' or 'inspect manually' cases) at will,
and we should not rely on convenient stochastic sampling assumptions or numerical representations of the cases in the flow.
Since the presented information volumes of information are presented at an enormous frequency,
 it is not even clear that one can do anything meaningful {\em at all}.

Examples of this settings are however ubiquitous in the modern digital society:
\begin{itemize}
\item In social media, we might want to make a detector of {\em hate-speech}, 
	based on the subjective judgement of a text message as {\em hate-speech} that a human expert can make.
	We are given a human expert who can look into a small number of text messages posted on a social media platform.
	Then we are challenged to make an algorithm that learns from those few examples so as to filter the many future cases for similar hate speech.
	
\item In the context of FinTech, we might want to make a Transaction Monitoring System (TMS) for fraud detection.
	Given a stream of financial transactions, each moving a certain amount between two (private) parties.
	Can we make an algorithm assisting our team of human experts to filter out fraudulent transactions from the high-frequency streams flowing through modern TMSs?
	
\end{itemize}
There is considerable work on detection in general, and fraud/fault/anomaly detection specifically:
we point 
to \cite{wald1947sequential,chernoff1972sequential} and citations for stochastic approaches to anomaly detection,
to \cite{benveniste1984detection} and citations for an analysis based on mathematical systems theory,
to \cite{van2004detection} and citations for approaches in statistical signal processing,
and to \cite{chandola2009anomaly} and citations for a recent survey based on methods of machine learning.
Most techniques that near the requirements are based on thresholds, need manual tuning, or make stringent assumptions about the nature of the data. 
In \cite{fado2016}, we present an algorithm named FADO that works explicitly with numerical representations of the transactions in the flow,
but is otherwise free from (stochastic) assumptions. The present work however is novel in adopting 
the expert framework \cite{cesabianchi2006,shalev2014understanding} for this task.
The expert framework allows one to work with (existing) expert systems expressed in Fortran77, Cobol, assembler, Mandarin Chinese, Urdu or other languages.
The aim of this paper is not to survey existing approaches\footnote{
Efforts as \url{https://www.kaggle.com/c/ieee-fraud-detection} aim for such benchmarking.}, 
but to indicate how a simple and intuitive algorithm addresses the challenge well.

\begin{figure}[htbp] 
   \centering
   \includegraphics[width=6in]{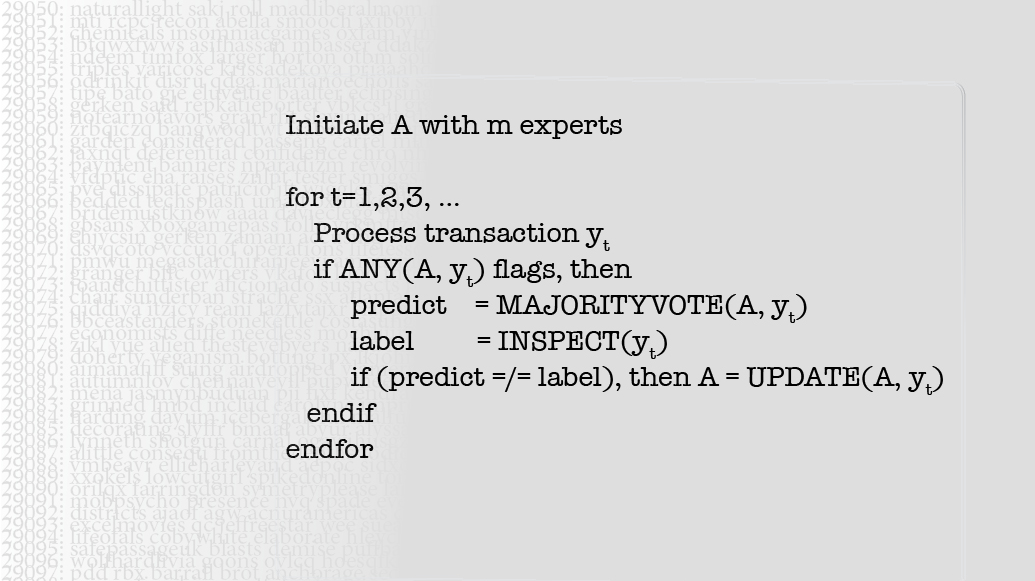} 
   \caption{Description of the HALFADO algorithm. Given are a stream of transactions (displayed in the background), and a set of $m$ experts.
   	Initially, all $m$ experts are in the active set $A$. When progressing 
	through the stream of transactions $\{y_t\}$, experts disappear from $A$ once implicated in a mistake of the majority vote of $A$.
	The algorithm demands a case to be inspected manually, only if at least one expert in the current $A$ flags the case.
   }
   \label{fig.halfado}
\end{figure}

We present in Fig. (\ref{fig.halfado}) an algorithm called HALFADO, extending the classical HALVING algorithm for classification (e.g. \cite{cesabianchi2006}, p.4).
HALFADO starts with a large number $m$ of experts: 
each expert can either flag a case as {\em suspicious}, or judge it as {\em normal}.
HALFADO maintains throughout the flow an active set of experts that proved earlier to perform well: 
the inadequate experts are dropped from the active set when progressing through the stream.
Assuming that a perfect expert exists amongst those in the initial active set  (i.e. the {\em existence assumption}), one can bound the number of such reductions.
By implementing the democratic majority vote on the current active set, a wrong evaluation (or {\em mistake}) will leave ('HALF') the active set with the correct minority.
In this way, we can at most make $\lfloor\log_2(m)\rfloor$ mistakes.

Finally - and most crucially - the algorithm flags a transaction as {\em potentially suspicious} if at least one expert in the active set flags the case as such.
When this happens, a human expert has to inspect manually the case (`Is the case really as {\em suspicious} as this expert says, or not?').
This allows HALFADO to learn from those manual inspections.
The aim is then to keep the number of such manual inspections to a minimum (classifying HALFADO as a semi-supervised learning algorithm), 
while not implicating the recall rate of the detector.
When the active set gets sufficiently small (i.e., is reduced sufficiently well), 
HALFADO ensures that  {\em alerting as potentially suspicious} 
and  prediction as {\em suspicious} aligns well.

Appendix A.1 provides an in-depth discussion of conceptual and theoretical considerations of HALFADO.
Appendix A.2 presents extensions of HALFADO for doing away with the existence assumption.
Appendix A.3 presents an extension of HALFADO for managing risk factors.

Now let us summarise the findings of the two case studies (see Appendix A.4 and A.5 for 
details\footnote{Data and code for reproducing the experimental results is at \url{https://fado.life/halfado/}.}).
In the case of information flows on a social media platform, we flag messages as hate-speech 
with a much higher precision and recall than what is achieved with a single-word detector. We are able to process 
$> 5,000$ text messages per second, and prompt `our' human experts only to look into 10,5\% of the traffic.
Note that we are not allowed to release the data publicly, but we provide code for collecting in realtime messages posted on a social media platform. 

In the FinTech case study for retail banking, we are interested in finding plausible risk factors that might lead to fraud.
After a while, HALFADO works with a small number of such risk factors. 
Overall, HALFADO requires one to manually inspect a mere 7\% of the presented financial transactions.
By comparison, one would have to investigate manually 0.05\% of the traffic if  {\em  one were to know the actual risk factors up front}.
This $6.95\%$ can then be seen as the price for not knowing (or having to learn) the risk factors.
Moreover, the experiment can handle almost $>40,000$ transactions {\em per second} on modest computational hardware.
The approach inherits accountability, privacy and ethical properties of the deployed experts.

In conclusion, HALFADO is theoretically proved and empirically seen to perform efficiently and computationally fast.
The expert-setting allows HALFADO to handle non-numerically represented data 
avoiding cumbersome feature design \cite{guyon2008feature} 
and boosting existing expert systems.
The conceptual ease of the approach encourages useful practical modifications: 
we found especially the extension towards 
managing risk factors to be thought provoking. 

We think that the natural ability of HALFADO to handle an unknown number of complimentary experts 
- each one representing potentially a different Root Cause \cite{julisch2003clustering} - points to an intriguing feature of this approach.
Another fundamental property is that {\em learning} here is really seen as {\em eliminating bad solutions}, 
rather than the approach to learning as optimising a cost/loss function.
This is reflected in the affluence of the method for handling conventional obstacles to learning as {\em local minima} or {\em overfitting} \cite{hastie01}.
It is however 
the practical relevance of this feather-light scheme,
its ability to work with existing expert systems, 
as well as the relief for the need for labeling, 
that is likely to power many more surprising applications.

\newpage
 \bibliographystyle{abbrv}

\newpage
\section*{Acknowledgements}

The research of KP is funded by the division of SysCon at UU/IT.
Part of the work was funded by a research grant from VR entitled `Adaptive Modelling and Online Learning with Large Sets of Unknowns', (2013-2017).

\section*{Author contributions}

KP is responsible for the conceptual investigations, theoretical derivations, as well as for writing the manuscript.
AB and MA have assisted to the result of the two case studies.

\section*{Author information}

KP (corresponding author, \verb|kp@it.uu.se|) 
is an associate professor ('docent') at Uppsala University (SE), department of Information Technology (IT), division of Systems and Control (SysCon).
MA and AB are master students at Uppsala University, department of Information Technology (IT).

\newpage
\appendix
\section{Supplementary material}

\subsection{Discussion of HALFADO.}

We present a conceptual and theoretical discussion of HALFADO.
\begin{itemize}
\item The information flow can be arbitrary. Specifically, we do not need to make assumptions on the stochastic sampling scheme nor 
	even the numerical representation of an individual transaction in the stream. 
	This renders the results practical even in cases where such assumptions are not even remotely valid.
\item We work explicitly with (a collection of) experts, and make sure that performance is almost as good 
	as the best one amongst them (i.e. the so-called {\em regret}). 
	The design of accurate experts is highly case-specific: the expert-framework allows us to abstract this away.
	In many cases, we start with enough experts so that {\em a perfect one} is likely to be amongst them.
\item The approach works explicitly with data streams, and does not have to store individual transactions.
	This alleviates privacy properties, or concerns prompted by the General Data Protection Regulation (GDPR) or the California Consumer Privacy Act (CCPA).
	Moreover, transparency, accountability and explainability properties are directly inherited from the used experts.
\item The approach is related to the class of boosting \cite{schapire1998boosting} and racing \cite{maron1997racing} algorithms.
	The extension to risk factors (as described in Appendix A.3) is based on insights on probability ('risk') in \cite{shafer2001probability},
	and has correspondence to the approach used in micro auctions in ad markets, as described in e.g. \cite{muthukrishnan2009}.
\end{itemize}
Theoretically, HALFADO comes with strong guarantees under 
the {\em existence assumption}.
\begin{itemize}
\item HALFADO with a majority vote can make at most $\lfloor\log_2(m)\rfloor$ mistakes.
	When the active set  got sufficiently small, this implies also a bounded number of false alarms.
\item The {\em existence assumption} ensures that there is always at least one expert that alarms 
	on a suspicious transaction, implying {\em total recall}.
\item The majority vote ($\theta=50\%$) can be replaced by a voting scheme using a smaller fraction (say, $\theta=1\%$).
	The number of mistakes is then bound as $\lfloor\log_{\frac{1}{1-\theta}}(m)\rfloor$.
\end{itemize}

\newpage
\subsection{Beyond the {\em existence assumption}.}

Existence of a perfect expert in the initial active set (the socalled {\em existence assumption}) is known to be stringent and unrealistic in many cases.
Can HALFADO still be used when the {\em existence assumption} doesnot hold (the socalled {\em agnostic} case)?
It turns out that modifications are needed - implicating theoretical, computational and practical elegance of the results.
Two different approaches are outlined.

The first approach - aligned to the literature on the topic - is as follows.
Rather than managing an active set of experts, the different experts are {\em weighted}.
The majority vote is adapted accordingly. 
Everytime an expert is implicated in a mistake, the corresponding weight decreases.
Hence, a single error is not final for an expert.
Down-weighting can e.g. be done exponentially, resulting in an EXP-like algorithm as in \cite{cesabianchi2006,mourtada2019optimality}.
However, such weighting-scheme implies that all experts need to be evaluated for each iteration of the algorithm, 
limiting computation speed (we would only be able to process a few instances per second on modest hardware).

We however follow a more practical scheme in the simulations.
Rather than implementing the above weight-based scheme for handling the agnostic case, 
the HALFADO as described before is made robust towards a small number of errors.
Once an expert is implicated in a mistake, it gets evicted now from the active set {\em with a certain probability $0<\alpha<1$}.
By choosing this $\alpha$ sufficiently low, a few mistakes will likely be tolerated.
This extension affects the bound to the regret only by adjusting the exponent and by allowing a small probability that results are unreliable.

The probability that an expert gets evicted from the active set after having seen exactly $n$ errors equals $(1-\alpha)^{n-1}  \alpha$.
The expected number of mistakes that an expert can tolerate (before this expert gets evicted from the active set) then becomes:
\begin{equation} 
	\sum_{i=0}^\infty i (1-\alpha)^{i-1}  \alpha 
	=
	\frac{\alpha}{1-\alpha} \sum_{i=1}^\infty i (1-\alpha)^i 
	=
	\frac{\alpha}{1-\alpha} \ \frac{1-\alpha}{\alpha^2}
	= 
	\frac{1}{\alpha}.
	\label{eq.prob2}
\end{equation}
Indeed, if $\alpha=1$, an expert will be evicted from the active set with its first mistake.
By lowering $\alpha$, experts are expected to withstand more (i.e. $1/\alpha-1$) mistakes.

\newpage
\subsection{Managing risk factors with HALFADO.}

Detecting fraud is often an Herculean challenge, and one cannot expect a single solution to perform well.
That lead us to shift gears, and focus on managing accurate risk factors instead.
Here, we explicitly acknowledge that the provided information is insufficient to judge wether a case with such increased risk is actual fraud or not.
For example, an international transaction in a TMS between two politically exposed persons (PEPs) might in 1\% of the cases turn out to be actual fraud:
a transaction between two PEPs is hence not necessarily fraudulent but warrants further (manual) investigation.
In other words: the rules do not claim to be causal but content themselves to be indicative.
In this example, one would have to inspect 100 financial transactions between two international PEPs to get on average a single fraud.
This reasoning is paramount in order to handle the low rates of fraudulent activities observed in modern TMS systems.

Then we modify HALFADO as follows.
Again, we work with an active set which contains initially a set of $m$ candidate risk factors.
When progressing through the stream of transaction, inadequate experts drop from this active set.
Rather than using the majority-vote and consequent HALVING, we use ideas common in economic (micro) auctions.
At each point, we let all the experts in the active set make a bid.
The expert with the largest bid (`wins the auction'), is granted access to the transaction.
If the transaction then turns out to be actual fraud, this expert gets a unit reward.
If it does not, the expert looses his investment.
So it is in the expert's own interest not to overestimate the risk of fraud, as otherwise he will be burdened with a large deficit by winning too many blank auctions.
If the deficit becomes too large, the expert will drop from the active set ({\em `go bankrupt'}).
On the other hand, if an expert underestimates the likelihood of gain, it will never win the bid.
Those two forces promote experts that make realistic investments, and weeds out the ones that get it consistently wrong.

Specifically, the following rule is used for managing the active set.
At any time $t$, the active set includes expert $i$ that satisfied at all previous times the constraint
\begin{equation}
	V_i^t \leq  P_i^t + c\sqrt{n_i^t\log_2(m)},
	\label{eq.limit}
\end{equation}
where $n_i^t\geq0$ denotes the number of times before $t$ where expert $i$ {\em won} the bid,
$P_i^t\geq0$ (the Profit) denotes the number of times before $t$ where expert $i$ {\em wan the bid and the corresponding transactions turned out to fraudulent},
and $V_i^t$ (the inVestment) denotes the {\em sum of investments that made expert $i$ win bids} before $t$.
Here $c>0$ is a constant (in the example below, we use $c=0.2$).
This rule is motivated by research on the Upper Confidence bound (UCB) class of algorithms  for the Multi-Armed bandit setting (see \cite{auer2002} and citations).

\newpage
\subsection{Case 1: Detecting hate speech in social media.}

This appendix reports results of making a detector for hate-speech\footnote{All results can be reproduced with the 
resources available at \url{https://fado.life/halfado}.}
based on text messages collected from a social media platform. The judgement of a text message as {\em hate-speech} is case-specific or even subjective.
HALFADO addresses this issue by working with a few annotated ('labelled') samples, and then to extrapolate this to the new messages.

Since we cannot do extended experimentation using the manual labeling scheme (Asking us: 'Is message $y_t$ indeed hate speech?),
we use an autonomous inspection rule checking wether the word 'hate' or 'kill' is included in this message. 
HALFADO has no knowledge of this rule, but tries to learn as well as possible from presented answers.

We collected $n=1,026,472$ text messages of maximum length 140 characters, from a popular social media platform.
We start HALFADO with $m=100,000$ experts, each based on the sign of a random projection of a Bag-of-Words  representation 
(BoW, \cite{manning1999foundations}) of size $500$.
Then amongst those experts, one is likely to find one that aligns well with the examples of {\em hate-speech} as collected during the process.
This results in a detection system able to process all messages in 173 seconds (for an average of processing 5,909 messages per second) 
on modest hardware  (MacBook Pro 2016, 3.1GHz Intel Core i5, 16Gb RAM).

After working through all messages, HALFADO results in 10 (5-18) experts that together implement a detection system.
This system prompts us to investigate 10.7\% (4.2\% - 21.3\%) of the traffic manually. This 89.3\% reduction translates in significant saving of resources.

\begin{figure}[htbp] 
   \centering
   \includegraphics[width=5in]{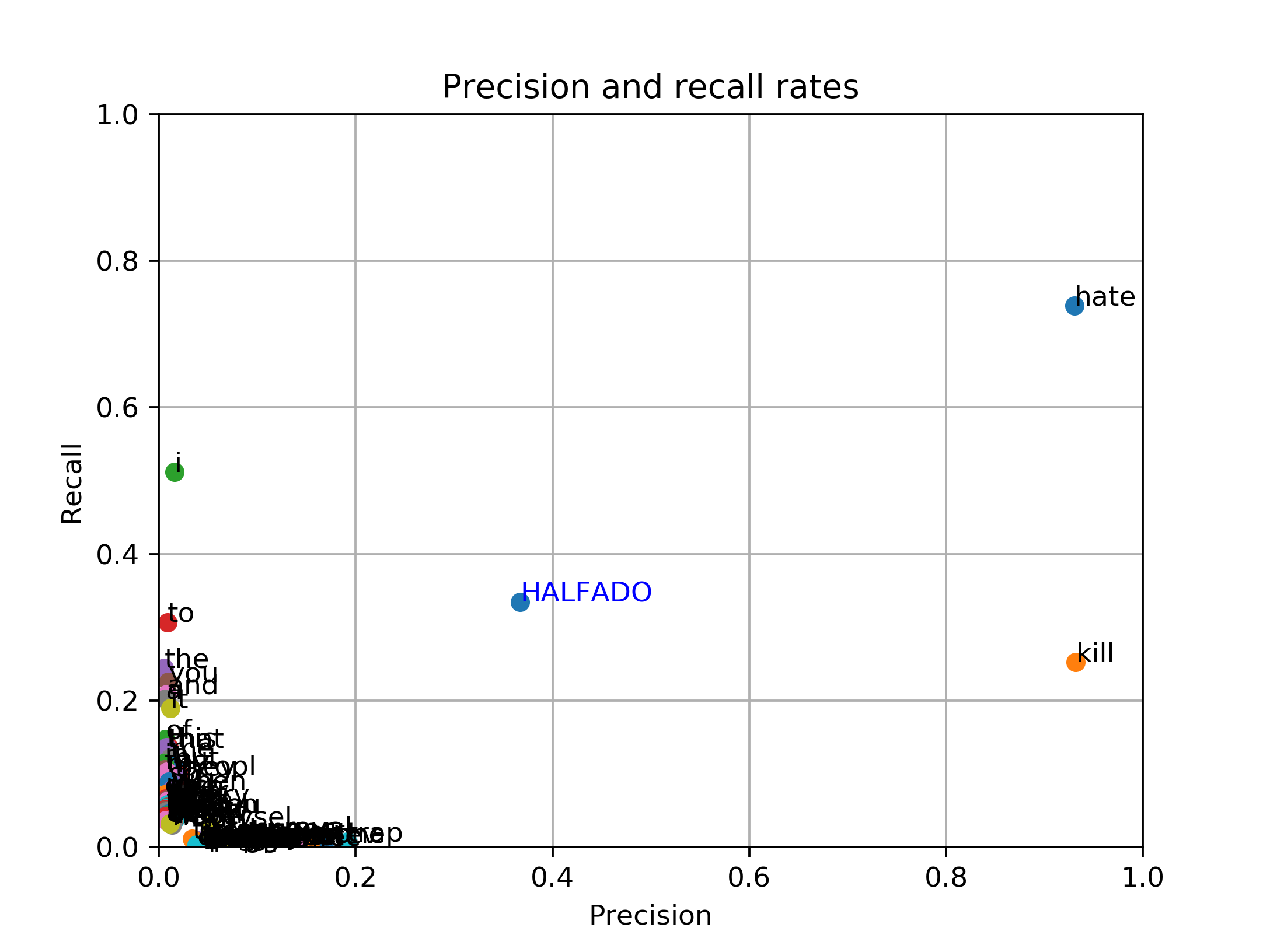} 
   \caption{Recall and precision rates of single-word-detectors and obtained by HALFADO.
   	HALFADO records a substantially higher recall (i.e. 33.5\%) for reasonable precision, than what is possible with a single-word-detector.
	Observe that in this experiment, we proxy manual labeling of examples as hate speech by using the presence of the word 'hate' or 'kill' in the message, 
	explaining their privileged position in this plot.
    }
   \label{fig.recall}
\end{figure}

We do not ensure that the existence condition is satisfied, allowing for non-total recall.
We however record a 33.5\% recall, which is substantial better than the recall achieved with a single-word-detector.
An example of a single-word-detector is 'This text contains the word \verb|molotov|', and would typically have high precision but low recall.
Figure  (\ref{fig.recall}) displays results.

\newpage
\subsection{Case 2: Detecting fraud in TMS in FinTech.}

This appendix reports results of the case study of a TMS in the context of retail banking 
(FinTech)\footnote{All results can be reproduced with the resources available at \url{https://fado.life/halfado/}.} .
For privacy reasons, we work with a. collection of 1,000,000 artificially generated financial transactions.
The following two risk factors are implemented on those: 
'Transactions moving from LV to DE' were with 10\% fraudulent', 
and 
'traffic shipping from BE to IT were in 5\% fraudulent\footnote{ These rules do not reflect reality beyond presented use.}.
Other transactions were not considered to be fraudulent.
this resulted in 30 fraudulent transactions in the set of 1,000,000 ones.
Can we perform as well without knowing those risk factors?

\begin{figure}[htbp] 
   \centering
   \includegraphics[width=6in]{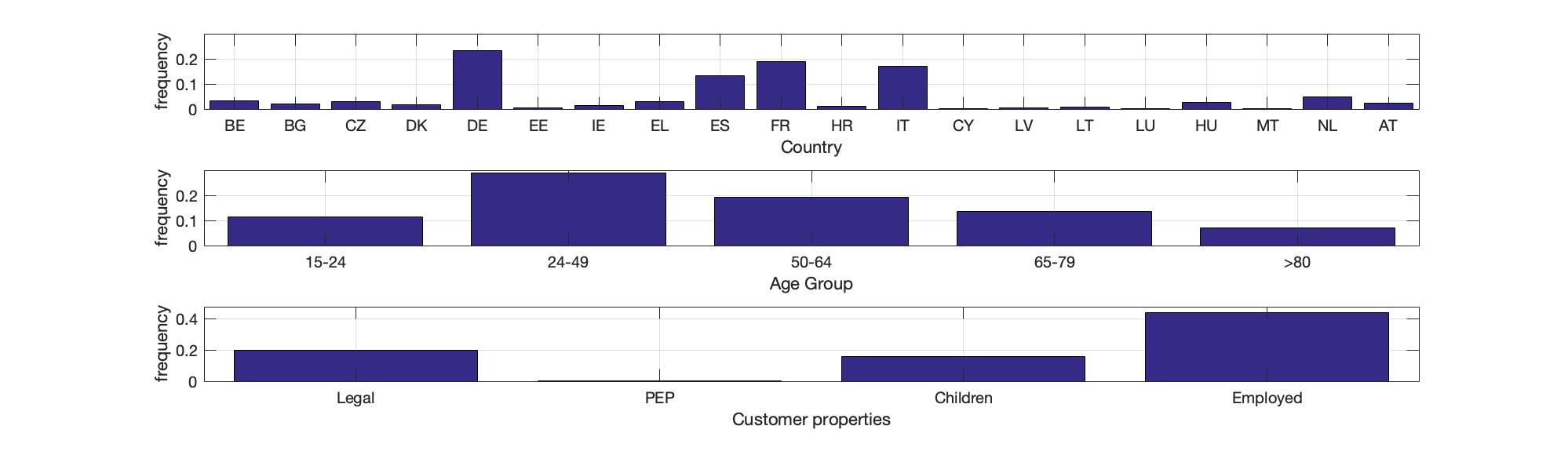} 
   \caption{\em 
   	The customers have properties distributed as indicated on the graphs (a frequency of 100\%=1 means that all customers characterize as positive to this feature).}
   \label{fig.cust}
\end{figure}

\begin{figure}[htbp] 
   \centering
   \includegraphics[width=6in]{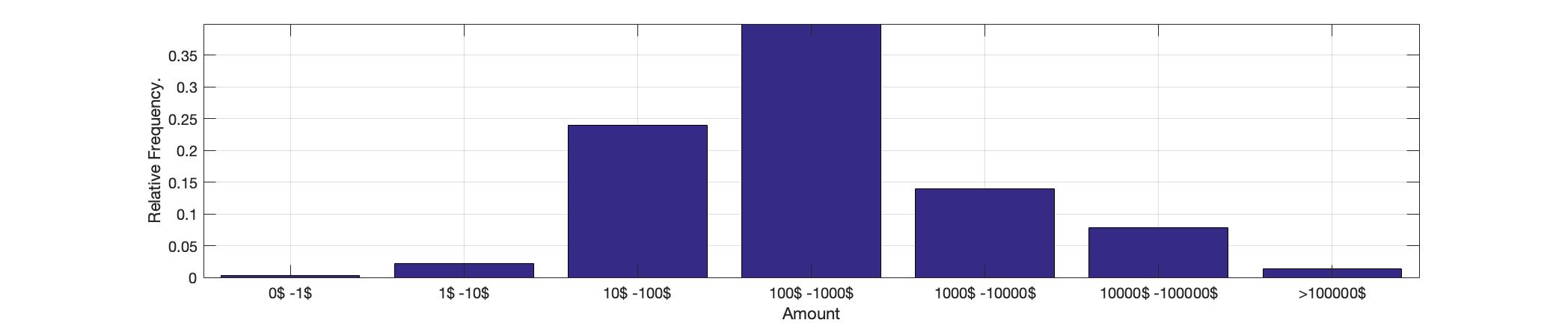} 
   \caption{\em The distribution of the amount of the 1,000,000 transactions. 
   Most transactions ship an amount of about 100\$.}
   \label{fig.tx}
\end{figure}

Each transaction ships an \verb|amount| from a sender to a receiver.
Such actor is characterised by 
\verb|country| of residence (we use the 20 most frequent countries in this example),
a \verb|PEP| (Politically Exposed Person) indicator,
a \verb|legal| indicator (wether the actor is a person or a legal entity),
and the age of the actor (divided in 5 age groups, as in Panel b of figure (\ref{fig.cust})),
wether he/she/it has children, and is employed at the time.
Figure (\ref{fig.tx}) displays how the amounts of the transactions are distributed.

\begin{figure}[htbp] 
   \centering
   \includegraphics[width=3in]{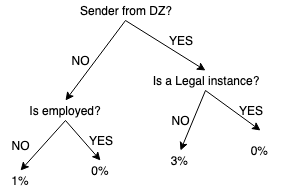} 
   \caption{\em An example of a decision tree of depth 2.
   	This tree encodes that 1\% of the traffic not from 'DZ' to an unemployed person, as well as 
	3\% of the traffic from 'DZ' to a non-legal receiver, is fraudulent. 
   }
   \label{fig.dt}
\end{figure}

The following results are obtained from repeatedly executing the example scripts.
We start with $m=1000$ experts, each encoding a randomly instantiated decision tree of depth 2 (see Fig. (\ref{fig.dt}) for an example of such tree).
The actual risk factors were ensured to be amongst them.
Then we apply HALFADO for managing risk factors.
We use a threshold of $0.2*\sqrt{n_i^t\log(m)}$ for eliminating experts that consistently overestimate risk (and end up hence with a large deficit).

On this laptop (MacBook Pro 2016, 3.1GHz Intel Core i5, 16Gb RAM), it took 14.9 seconds to process all 1,000,000 transactions
(processing on average 68,706 transactions per second).
The algorithm asked us to inspect manually 7\% (3.3\% - 17.9\%) of the presented transactions.
If the actual rules were known up front, then one would have to investigate `an optimal' 0.05\% of the transactions. 
We find that HALFADO reduces the set of $m=1000$ experts to 3 (2-5) plausible experts:
those include in most (87.1\%) runs both actual rules 
(i.e., '10\% of the traffic from LV to DE', and '5\% of the traffic from BE to IT', is fraudulent) encoded in the example.
The expert corresponding to the rule '10\% of the traffic from LV to DE'  won the auction 50 (44-58) times, making a total investment of 5\$.
On average, it was granted a 1\$ profit for 5 (4-7) times as the investigated transaction proved indeed to be fraud.
Note that the logic of the algorithm ensures a total recall, but that precision is typically very low.

Considering the computational ease, this result is beyond expectation.
However, the observed rates of inspection (7\% of 1,000,000 transactions amounts to 70,000 manual inspections)  
suggest to adopt a multi-level approach for handling followup.

\end{document}